\pdfoutput=1

\documentclass[11pt]{article}

\newif\ifarxiv
\arxivtrue

\ifarxiv
    \usepackage{acl}
\else
    \usepackage[review]{acl}
\fi

\usepackage{times}
\usepackage{latexsym}
\usepackage{amsmath,amssymb,amsthm}

\usepackage{bbm}
\usepackage{mathabx,mathrsfs}
\usepackage{relsize}
\PassOptionsToPackage{hyphens}{url}\usepackage{hyperref}
\usepackage{xurl}
\usepackage[T1]{fontenc}

\usepackage[utf8]{inputenc}
\usepackage{microtype}
\usepackage{CJKutf8}
\usepackage{bm}
\usepackage{array}
\usepackage{booktabs}
\usepackage[countmax]{subfloat}
\usepackage{color}
\usepackage{xcolor,colortbl}
\usepackage{mleftright}
\usepackage{caption}
\usepackage{subcaption}
\usepackage{pifont}
\usepackage{multirow}
\usepackage{hyperref}
\usepackage{paralist}
\usepackage{bigstrut}
\usepackage{microtype}
\usepackage{tikz}
\usepackage{pgfplots}
\usepackage{tikz-dependency}
\usepackage{graphicx}
\usepackage{tcolorbox}
\usepackage{natbib}
\usepackage{tikzsymbols}
\usepackage{scalerel}
\usepackage{svg}
\usepackage{adjustbox}
\usepackage{tikz-qtree}
\usepackage{nicematrix}
\usepgfplotslibrary{groupplots}
\usepgfplotslibrary{fillbetween}
\usepgfplotslibrary{statistics}
\usetikzlibrary{fit, calc, plotmarks, decorations.pathreplacing, decorations.markings, positioning, arrows.meta, shapes.geometric, backgrounds, graphs}
\pgfplotsset{compat=1.18}
\usepackage{float}

\usepackage{graphicx}

\pgfdeclarelayer{background}  
\pgfsetlayers{background,main}  

\usepackage{todonotes}
\usepackage{marginnote}
\usepackage{makecell}
\newcommand{\hochkomma}{$^{,}$}
\newcommand{\note}[4][]{{
  \todo[author=#2,color=#3,size=\footnotesize,fancyline,caption={},#1]{#4}
}}

\definecolor{tticblue}{RGB}{0, 94, 184}

\newcommand{\zhqiao}[2][]{{\note[#1]{zhqiao}{tticblue!20}{#2}}}
\newcommand{\Zhqiao}[2][]{\zhqiao[inline,#1]{#2}\noindent}

\def\hwemoji@insert#1{\scalerel*{\includegraphics[page=#1]{assets/hwemoji-assets.pdf}}{X}}
\newcommand\orange{1}
\newcommand\coconut{2}
\newcommand\mailemoji{\hwemoji@insert{3610}}
\newcommand\Letter{{\fontfamily{mvs}\fontencoding{U}\selectfont\char66}}

\usepackage{tabularx}

\definecolor{yanhong}{RGB}{130, 17, 31}
\definecolor{yanlan}{RGB}{20, 74, 116}
\definecolor{json_blue}{RGB}{15, 89, 164}
\definecolor{json_red}{RGB}{192, 44, 53}
\definecolor{figure_green}{RGB}{32, 137, 77}
\definecolor{figure_blue}{RGB}{52, 108, 156}
\definecolor{figure_red}{RGB}{192, 44, 53}
\definecolor{figure_orange}{RGB}{250, 126, 35}
\definecolor{table_blue}{RGB}{92, 179, 204}
\definecolor{table_red}{RGB}{238, 63, 77}
\definecolor{table_orange}{RGB}{250, 126, 35}
\definecolor{figure_light_green}{RGB}{198, 223, 200}
\definecolor{figure_light_blue}{RGB}{208, 223, 230}
\definecolor{figure_light_red}{RGB}{192, 44, 53}
\definecolor{figure_light_gray}{RGB}{220, 220, 220}
\definecolor{figure_gray}{RGB}{160, 160, 160}

\newcommand\wz{\phantom{0}}

\newcommand\bgood{\hspace{-0.05em}\rlap{\scalebox{0.9}{$^\upharpoonright$}}}
\newcommand\sgood{\hspace{-0.15em}\rlap{\scalebox{0.9}{$^\downharpoonright$}}}
\newcommand\correct[1]{\textcolor{figure_blue}{{#1}}}
\newcommand\wrong[1]{\textcolor{figure_red}{{{#1}}}}

\newcommand \footnotetextonly[1]
{
    \let \backupfootnote \thefootnote
    \let \thefootnote \relax
    \footnotetext{#1}
    \let \thefootnote \backupfootnote
    \let \backupfootnote \imreallyundefinedcommand
}

\title{DISC: Plug-and-Play Decoding Intervention with Similarity of Characters \\ for Chinese Spelling Check}

\author{Ziheng Qiao\rlap{$^{\orange}$},\ \ Houquan Zhou\rlap{$^{\orange}$},\ \ Yumeng Liu\rlap{$^{\orange}$},\ \ Zhenghua Li\rlap{$^{\orange\text{\Letter}}$},\ \ \ \  Min Zhang$^{\orange}$\ \ \\
{\bf  Bo Zhang\rlap{$^{\coconut}$},\ \ Chen Li\rlap{$^{\coconut}$},\ \ Ji Zhang\rlap{$^{\coconut}$},\ \  Fei Huang\rlap{$^{\coconut}$}}  \\%
$^{\orange}$School of Computer Science and Technology, 
Soochow University, China \\%
$^{\coconut}$DAMO Academy, Alibaba Group, China\\%
\texttt{\{zhqiao,hqzhou,ymliu14\}@stu.suda.edu.cn},\\%
\texttt{\{zhli13,minzhang\}@suda.edu.cn},\\%
\texttt{\{klayzhang.zb,puji.lc,zj122146,f.huang\}@alibaba-inc.com}}

\begin{document}
\maketitle
\begin{CJK*}{UTF8}{gkai}
    \ifarxiv%
        \footnotetextonly{\!\!\Letter\ Zhenghua Li is the corresponding author.}
    \fi%
    \begin{abstract}

One key characteristic of the Chinese spelling check (CSC) task is that incorrect characters are usually similar to the correct ones in either phonetics or glyph. 
To accommodate this, previous works usually leverage confusion sets, which suffer from two problems, i.e., difficulty in determining which character pairs to include and lack of probabilities to distinguish items in the set. 
In this paper, we propose a light-weight plug-and-play DISC (i.e., decoding intervention with similarity of characters) module for CSC models.
DISC measures phonetic and glyph similarities between characters and incorporates this similarity information only during the inference phase.
This method can be easily integrated into various existing CSC models, such as ReaLiSe, SCOPE, and ReLM, without additional training costs.
Experiments on three CSC benchmarks demonstrate that our proposed method significantly improves model performance, approaching and even surpassing the current state-of-the-art models.

\end{abstract}

    \section{Introduction}
\label{intro}

Given an input sentence,  Chinese spelling check (CSC) aims to detect incorrect characters and modify each into a correct character \cite{yu-li-2014-chinese,Xu-etal-2021-realise}. Table \ref{tab:examples_of_introduction} gives two examples. 
Spelling errors degrade reading efficiency, and sometimes even lead to misunderstanding. 
The authority or attitude of the writer may be doubted if their document contains simple spelling errors.  
Moreover, spelling errors substantially hurt the performance of subsequent NLP models.  


\begin{table}[t!]
    \centering
    {
            \resizebox{0.48\textwidth}{!}{%
            \begin{NiceTabular}{ll}
                \toprule
                \Block{2-1}{Input} & 记得戴眼\wrong{睛(jīng)}。 \bigstrut\\
                                  & Remember to wear \wrong{eyes}. \\
                \hdottedline
                \Block{2-1}{Reference} & 记得戴眼\correct{镜(jìng)}。\bigstrut \\
                                      & Remember to wear \correct{glasses}. \\
                \midrule
                \midrule
                \Block{2-1}{Input} & 从商场的\wrong{人(rén)}口进去。\bigstrut \\
                                  & Enter through the mall's \wrong{population}. \\
                \hdottedline
                \Block{2-1}{Reference} & 从商场的\correct{入(rù)}口进去。\bigstrut \\
                                      & Enter through the mall's \correct{entrance}. \\
                \bottomrule
            \end{NiceTabular}}
    }
    \caption{
        Two CSC examples. 
        ``睛''(jīng, eyes) and ``镜''(jìng, glasses) are a pair of characters that are similar in phonetics, and ``人''(human) and ``入''(enter) are similar in glyph.
    }
    \label{tab:examples_of_introduction}
\end{table}

As is well known, spelling errors in Chinese texts have three major sources, i.e., 1) from keyboard typing with some input methods, 2) from image or document scanning with some optical character recognition (OCR) software, and 3) from speech-to-text translation with some automatic speech recognition (ASR) software. 
Nowadays, most Chinese users employ Pinyin-based input methods. 
Considering the three sources, we can see that the incorrect character in most cases is similar to the underlying correct one in phonetics or glyph, sometimes in both. 
This is a key characteristic of the CSC task.

Previous works employ confusion sets to leverage such similarities among characters \cite{yeh-etal-2013-chinese,huang-etal-2014-chinese,xie-etal-2015-chinese,Cheng-etal-2020-SpellGCN,Huang-etal-2023-module}. 
Formally, a confusion set is denoted as 
$\mathcal{C} = \{(c^{1}_i, c^{2}_i)\}_{i=1}^{M}$, where each pair 
$(c^{1}_i, c^{2}_i)$ represents a pair of characters and means that $c^{1}_i$ may be mistakenly replaced by $c^{2}_i$ in real texts. 

As a representative work, \citet{wang-etal-2018-hybrid} construct a confusion set via two channels. First, they add noise into glyph images and apply OCR. Second, they apply ASR to parallel speech/text data. 
Their confusion set covers about 5K characters and consists of 19K character pairs that are likely to be confused with each other in written texts. 


The most direct and popular use of confusion sets is to constrain the search space during the inference phase. The model can only consider character pairs in $\mathcal{C}$. More specifically, if $(c_1, c_2) \notin \mathcal{C}$, the model can never change $c_2$ into $c_1$. 
The justification for such \emph{constrained decoding} is that the resulting sentence may deviate from the meaning of the input sentence (i.e., unfaithfulness), if the model replaces a character with a totally unrelated new one. 



Despite their popularity and usefulness, confusion sets have two problems.  First, it is difficult to set criteria to decide the inclusion or exclusion of certain character pairs. This renders the construction of confusion sets 
highly empirical, sometimes requiring manual intervention. 
Second, there is no probability to distinguish which character pairs are more likely to be confused than others in $\mathcal{C}$. 

\begin{figure*}[th]
    \centering
    \includegraphics[width=\textwidth, trim=0cm 0cm 0cm 0cm, clip]{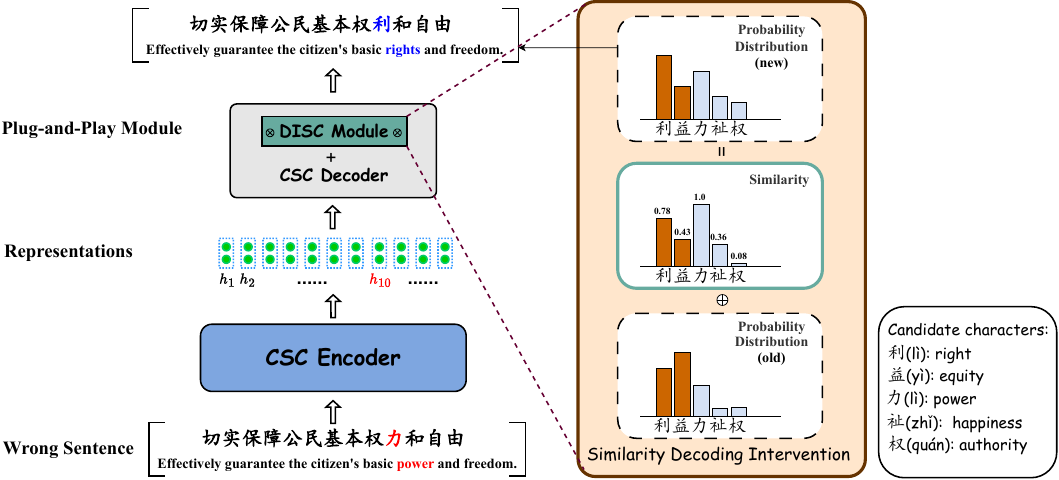} 
    \caption{Overview of DISC. It intervenes in the CSC decoder with the similarity between the potential error character and its candidate characters. The DISC module intervenes in the probability distribution results of the CSC model based on specific similarity, favoring the selection of more similar confusing characters.}
    \label{fig:big_picture}
\end{figure*}

As a replacement for confusion sets, we propose a lightweight plug-and-play DISC (\textbf{d}ecoding \textbf{i}ntervention with \textbf{s}imilarity of \textbf{c}haracters) module. DISC derives probability-based similarities among characters in both phonetics and glyph, and uses them to intervene in the decoding process. Similar to the confusion set, our DISC aims to enhance the model's precision. However, for datasets that lack or have no in-domain training data, DISC may result in under-corrections due to the model's conservative predictions, leading to unstable recall. To address this, we propose a copy-punishment solution to balance precision and recall.

It is worth noting that DISC is featured in compatibility. 
On the one hand, DISC is compatible with the ways to derive probabilities for representing character similarity. 
On the other hand, 
DISC is compatible with almost all the current mainstream CSC models, such as SoftMasked-BERT \cite{zhang-etal-2020-softmasked}, ReaLiSe \cite{Xu-etal-2021-realise}, SCOPE \cite{Li-etal-2022-SCOPE}, and ReLM \cite{Liu-etal-2024-ReLM}.

Experiments and analyses on popular benchmark datasets, i.e., SIGHANs, ECSpell, and LEMON, demonstrate that our DISC module can significantly enhance the error correction performance of CSC models.
This improvement does not require additional training costs and only slightly affects the decoding efficiency of the model.
We release our code at \url{https://github.com/zhqiao-nlp/DISC}.

    \section{The Basic CSC Model}
\label{sec:basic_approach:basic_csc_model}



Given an input sentence consisting of $n$ characters, denoted as $\boldsymbol{x}=x_1 x_2 \cdots x_n$, the goal of a CSC model is 
 to output a corresponding correct sentence, denoted as  $\boldsymbol{y}=y_1 y_2 \cdots y_n$, in which all erroneous characters in $\boldsymbol{x}$ are replaced with the correct ones. 

 Presently, mainstream approaches treat CSC as a character-wise classification problem  \cite{zhang-etal-2020-softmasked,Liu-etal-2021a-pretrain,Xu-etal-2021-realise}, i.e., determining whether a current character should be kept the same or be replaced with a new character. 

\paragraph{Encoding.} 
Given $\boldsymbol{x}$, the encoder of the CSC model generates representations for each character:
 \begin{equation}
    \begin{aligned}
        \boldsymbol{h}_1 \cdots \boldsymbol{h}_n = \mathbf{Encoder}(\boldsymbol{x}). 
        \label{eq:Encoder}
    \end{aligned}
\end{equation}

To leverage the power of pre-trained language models, a BERT-like encoder is usually employed. 


\paragraph{Classification.} 
For each character position, for instance $\boldsymbol{h}_i$, the CSC model employs MLP and softmax layers to obtain a probability distribution over the whole character vocabulary $\mathcal{V}$:
\begin{equation}
    \begin{aligned}
        p(y \mid \boldsymbol{x}, i) & = \texttt{softmax}(~ \texttt{MLP}(\boldsymbol{h}_i) ~)[y].
        \label{eq:Distribution}
    \end{aligned}
\end{equation}

During the evaluation phase, the model selects the character with the highest probability, i.e., $y^*  = \mathop{\arg\max}_{y \in \mathcal{V}} p(y \mid \boldsymbol{x}, i)$. 

\paragraph{Training.} The typical training procedure consists of 2–3 steps for the CSC task. First, automatically synthesize large-scale CSC training data by replacing some characters with others randomly, sometimes constrained by a given confusion set. 
Second, train the CSC model on the synthesized training data. 
Third, fine-tune the model on a small-scale in-domain training data, if the data is available. 

\section{Our Approach}

In this paper, we propose a simple plug-and-play module to intervene in the classification (or prediction) process of any off-the-shelf CSC model. The basic idea is to adjust the probability distribution according to the similarity between a candidate character $y$ and the original character $x_i$:
\begin{equation}
    \begin{aligned}
        \texttt{Score}(\boldsymbol{x}, i, y) & = p(y \mid \boldsymbol{x}, i) + \alpha \times \texttt{Sim}(x_i, y),
    \end{aligned}
        \label{eq:New Distribution}
\end{equation}
where $\texttt{Sim}(\cdot)$ gives the similarity between two characters, and $\alpha$ is a hyperparameter and we set $\alpha=1.1$ for all datasets and basic models according to a few preliminary experiments. 
We use $\texttt{Score}(\cdot)$ to denote the replacement likelihood since the value is no longer a probability.  

Our experiments show that by encouraging the model to prefer similar characters, our approach achieves a consistent and substantial performance boost on all CSC benchmark datasets.

We measure character similarity from two perspectives, i.e., phonetic and glyph:
\begin{equation}\label{eq:Similarity}
    \begin{aligned}
\texttt{Sim}(c_1, c_2)  =  \beta & \times \texttt{Sim}^\texttt{P}(c_1, c_2) \\ +  (1 - \beta) & \times \texttt{Sim}^\texttt{G}(c_1, c_2),
    \end{aligned}
\end{equation}
where $\beta$ is an interpolation hyperparameter, our experiments in Section \ref{sec:RobustnessofHyperparameters} demonstrate that the model achieves good and stable performance when it is set to 0.7.

\subsection{Phonetic Similarity}

Given two characters, we employ the pypinyin library to obtain the Pinyin sequences,\footnote{\url{https://pypi.org/project/pypinyin}} e.g., 
``忠'' (zhong) and ``仲'' (zhong),\footnote{We do not use the tone information, e.g., ``忠'' (zhōng) and ``仲'' (zhòng), which is not helpful for model performance according to our preliminary experiments. We suspect the reason is that Pinyin-based input methods do not require users to input the tones. Therefore, tones are not directly related to spelling errors.} 
and then compute the phonetic similarity based on the edit distance over their Pinyin sequences:
\begin{equation}
    \begin{aligned}
        \texttt{Sim}^\texttt{P}(c_1, c_2) & = 1 - \frac{\texttt{LD}(\texttt{py}(c_1), \texttt{py}(c_2))}{\texttt{len}(\texttt{py}(c_1) + \texttt{py}(c_2))},
        \label{eq:Pinyin Levenshtein Distance Similarity}
    \end{aligned}
\end{equation}
where $\texttt{LD}(\cdot)$ gives the Levenshtein distance,\footnote{Levenshtein distance is a type of edit distance. We set the weights of the three types of operations, i.e., deletion, insertion and substitutions, as 1/1/2 respectively. 
}
and $\texttt{len}(\cdot)$ gives the total length of the two sequences. 

\paragraph{Handling polyphonic characters.} Given two characters, we enumerate all possible Pinyin sequences of each character, and adopt the combination that leads to the highest similarity. 

We have also tried more sophisticated strategies. For instance, we follow \citet{yang-etal-2023-chinese} and give higher weights to certain phoneme (consonant or vowel) pairs, since they are more likely to cause spelling errors.  
However, our preliminary experiments show that our simple strategy in Eq.~\eqref{eq:Pinyin Levenshtein Distance Similarity} works quite robustly.

\subsection{Glyph Similarity}


According to \citet{liu-etal-2010-visually}, 83\% of Chinese spelling errors are related to pronunciation, while 48\% are with glyphs, indicating that a considerable proportion is related to both.  
Therefore, it is necessary to consider the glyph information when computing character similarity. 

Pinyin sequences can largely encode the phonetics of Chinese characters. In contrast, it is much more complex to represent character glyphs. 
In this work, we compute and fuse glyph similarity from four aspects:
\begin{equation}
    \begin{aligned}
        \texttt{Sim}^\texttt{G}(c_1, c_2) & = \frac{\sum_{i=1}^{4} \texttt{Sim}^{\texttt{G}}_i(c_1, c_2)}{4}.
        \label{eq:Glyph Similarity}
    \end{aligned}
\end{equation}

\paragraph{Four-corner code.}
The four-corner method is widely used in Chinese lexicography for indexing characters. Given a character, it gives four digits ranging from 0 to 9, corresponding to the shapes at the four corners of the character's glyph, respectively. For instance, the four-corner code is \texttt{5033} for ``忠'', and \texttt{2520} for ``仲''. 

Then, we use the digit-wise matching rate between two codes as the similarity:
\begin{equation}
    \begin{aligned}
        \texttt{Sim}^{\texttt{G}}_1(c_1, c_2) & = \frac{\sum_{i=1}^4{\mathbbm{1}(\texttt{FC}(c_1)[i]=\texttt{FC}(c_2)[i])}}{4},
        \label{eq:Glyph Similarity 1}
    \end{aligned}
\end{equation}
where $\texttt{FC}(\cdot)$ gives the four-digit code, and $\mathbbm{1}$ is the indicator function.

\paragraph{Structure-aware four-corner code.} 
One important feature of Chinese characters is that a complex character can usually be decomposed into simpler parts, and each part corresponds to a simpler character or a radical. 
Most radicals are semantically equivalent to some character, e.g., ``亻''  to ``人''. 

Such structural decomposition directly reveals how characters are visually similar to each other. 
Motivated by this observation, we design a structure-aware four-corner code for each character. For example, \\ 
$~~~~$``忠'': \texttt{C5000C3300} (``中'': \texttt{5000}; ``心'': \texttt{3300})\\
$~~~~$``仲'': \texttt{B8000B5000} (``人'': \texttt{8000}; ``中'': \texttt{5000})\\
where ``\texttt{C}'' leading a four-coner code means up-down structure, and ``\texttt{B}'' means left-right structure. 

Then we compute the similarity based on the Levenshtein distance as follows:
\begin{equation}
    \begin{aligned}
        \texttt{Sim}^\texttt{G}_2(c_1, c_2) & = 1 - \frac{\texttt{LD}(\texttt{SFC}(c_1), \texttt{SFC}(c_2))}{\texttt{len}(\texttt{SFC}(c_1) + \texttt{SFC}(c_2))},
        \label{eq:Glyph Similarity 2}
    \end{aligned}
\end{equation}
where $\texttt{SFC}(\cdot)$ gives the structure-aware code of a character. 

\paragraph{Stroke sequences.} 
Four-corner codes focus on the shapes of the four corners. Some very similar characters may obtain quite different codes, e.g.,  ``木'' ($\texttt{4090}$) vs. ``本'' ($\texttt{5023}$). 
To address this issue, we utilize stroke sequence information, which encodes how a character is handwritten stroke by stroke. For example, \\
\indent ``木'': 一丨ノ、\ (4 strokes)\\
\indent ``本'': 一丨ノ、一 \ (5 strokes)

Then we compute two similarity metrics from two complementary viewpoints. The first metric is based on Levenshtein distance:
\begin{equation}
    \begin{aligned}
        \texttt{Sim}^\texttt{G}_3(c_1, c_2) & = 1 - \frac{\texttt{LD}(\texttt{SS}(c_1), \texttt{SS}(c_2))}{\texttt{len}(\texttt{SS}(c_1) + \texttt{SS}(c_2))},
        \label{eq:Glyph Similarity 3}
    \end{aligned}
\end{equation}
where $\texttt{SS}(\cdot)$ gives the stroke sequence of a character. 

The second metric considers the longest common subsequence, i.e., $\texttt{LCS}(\cdot)$:
 \begin{equation}
    \begin{aligned}
        \texttt{Sim}^\texttt{G}_4(c_1, c_2) & = \frac{\texttt{LCS}(\texttt{SS}(c_1), \texttt{SS}(c_2))}
        {\max(\texttt{len}(\texttt{SS}(c_1)), \texttt{len}(\texttt{SS}(c_2)))}.
        \label{eq:Glyph Similarity 4}
    \end{aligned}
\end{equation}

According to Eq.~\eqref{eq:Similarity}, and supposing $\beta=0.7$, we get the similarity between ``忠'' and ``仲'' being:
\begin{equation*}
    0.7 \times 1 + 0.3 \times \frac{0 + 0.56 + 0.57 + 0.5}{4} = 0.82.
\end{equation*}
    
\section{Experimental Setup}
\begin{table*}[tb!]
    \renewcommand{\arraystretch}{0.95}
    \centering
    \scalebox{0.92}{
        \begin{NiceTabular}{lcccc|cccc|cccc}
            \toprule
            \rowcolor[gray]{1.0}
            \Block[l]{2-1}{\textbf{Models}} & \Block[c]{1-4}{\textbf{SIGHAN15}} & & & & \Block[c]{1-4}{\textbf{SIGHAN14}} & & & & \Block[c]{1-4}{\textbf{SIGHAN13}} & & & \\
            \rowcolor[gray]{1.0} & C-P\bgood & C-R\bgood & C-F\bgood & FPR\sgood & C-P\bgood & C-R\bgood & C-F\bgood & FPR\sgood & C-P\bgood & C-R\bgood & C-F\bgood & FPR\sgood \\
            \midrule
            \rowcolor[gray]{.95}
            \Block[c]{1-13}{{Previous SOTAs}} & & & & & & & & & & & & \\
            \Block[l]{1-1}{SpellGCN} & 72.1          & 77.7          & 75.9 & -- & 63.1          & 67.2          & 65.3 & -- & 78.3          & 72.7          & 75.4 & -- \\
            \Block[l]{1-1}{ReaLiSe} & 75.9          & 79.9          & 77.8                            & 12.0 & 66.3          & 70.0          & 68.1                            & 14.9 & 87.2          & 81.2          & 84.1                            & 10.3 \\
            \Block[l]{1-1}{SCOPE\rlap{$^\dagger$}} & 78.7          & 83.5          & 81.0                            &  11.3 & 67.1          & 71.2          & 69.5                            & 14.8 & 86.5          & 82.1          & 84.2                            & 17.2 \\
            \Block[l]{1-1}{SCOPE\;+\;DR-CSC} & \textbf{80.3} & 82.3          & 81.3                            & -- & \textbf{69.3}          & 72.3          & 70.7                            & -- & 87.7          & 83.0          & 85.3                            & -- \\
            \Block[l]{1-1}{ReLM\rlap{$^\dagger$}} & 76.8          & \textbf{83.9} & 80.2                            & 12.7 & 63.7          & 72.3          & 67.7                            & 17.5 & 85.0          & 82.3          & 83.7                            & 10.8 \\
            \midrule
            \rowcolor[gray]{.95}
            \Block[c]{1-13}{{LLMs Results}} & & & & & & & & & & & & \\
            \Block[l]{1-1}{{GPT3.5}} & 32.7 & 38.4          & 35.3                            & 33.8 & 39.7 & 22.1 & 28.4 & 14.6 & 57.1 & 27.1 & 36.7 & 13.8 \\
            \Block[l]{1-1}{{GPT4}} & 36.5 & 49.2          & 41.9                            & 40.8 & 32.8 & 45.0 & 38.0 & 43.5 & 47.3 & 45.7 & 46.5 & 44.8 \\
            \midrule
            \rowcolor[gray]{.95}
            \Block[c]{1-13}{{Ours}} & & & & & & & & & & & & \\
            \Block[l]{1-1}{ReaLiSe\;+\;DISC} & 77.0          & 79.9          & 78.4\rlap{$^\uparrow$}          & 11.3\rlap{$^\downarrow$} & 68.2          & 70.2          & 69.2\rlap{$^\uparrow$}          & \textbf{13.7}\rlap{$^\downarrow$} & 87.6          & 81.1          & 84.2\rlap{$^\uparrow$}          & 10.3 \\
            \Block[l]{1-1}{SCOPE\;+\;DISC} & 80.2          & 83.4          & \textbf{81.8}\rlap{$^\uparrow$}          & 10.0\rlap{$^\downarrow$} & \textbf{69.3}          & 72.5          & 70.9\rlap{$^\uparrow$}          & \textbf{13.7}\rlap{$^\downarrow$} & 88.0          & 83.0          & 85.4\rlap{$^\uparrow$}          & 17.2 \\
            \Block[l]{1-1}{ReLM\;+\;DISC} & 79.8          & 83.1          & 81.4\rlap{$^\uparrow$} & \textbf{\wz{9.5}}\rlap{$^\downarrow$} & 68.6 & \textbf{73.7} & \textbf{71.0}\rlap{$^\uparrow$} & 14.3\rlap{$^\downarrow$} & \textbf{88.4} & \textbf{83.3} & \textbf{85.8}\rlap{$^\uparrow$} & \textbf{\wz{7.6}}\rlap{$^\downarrow$} \\
            \bottomrule
        \end{NiceTabular}
    }
    \caption{
        Sentence-level performance on the SIGHAN13, SIGHAN14 and SIGHAN15 test sets. Precision ($\mathrm{P}$),
        recall ($\mathrm{R}$) and $\mathrm{F}_1$ for correction are reported (\%). Results marked with ``$\dagger$'' are obtained by reruning the official code released by \citet{Li-etal-2022-SCOPE} and \citet{Liu-etal-2024-ReLM}. Other baseline results are directly taken from their literature. Apart from SpellGCN, all models apply post-processing on SIGHAN13, which removes all detected and corrected ``地'' and ``得'' from the model output before evaluation. ``+ DISC'' means adding DISC module in the decoder. $\alpha$ and $\beta$ are assigned the values 1.1 and 0.7, respectively.
    }
    \label{tab:sighans_experiment}
\end{table*}


\subsection{Datasets}
Following the conventions of previous work, we employ the test sets of the SIGHAN 13/14/15 datasets \cite{wu-etal-2013-chinese,yu-etal-2014-overview,tseng-etal-2015-introduction} as our evaluation benchmarks.

However, many previous studies have pointed out that the SIGHAN datasets may not represent real-world CSC tasks, as they are derived from Chinese learner texts.
To address this limitation, we also conduct experiments on the ECSpell \cite{Lv-etal-2023-ECSpell} and LEMON \cite{Wu-etal-2023c-LEMON} datasets, which are derived from Chinese native-speaker (CNS) texts and encompass a wide range of domains. It is worth noting that LEMON does not have a dedicated training set, making it an excellent test set for evaluating a model's generalization ability. 
Due to space constraints, we selected results from four domains for display and provided the average performance across all seven domains.

The details of these datasets are in Appendix \ref{sec:appendix:datasets}.

\subsection{Baseline Models}
We select three representative BERT-style models as our baselines: \textbf{ReaLiSe}, \textbf{SCOPE}, and \textbf{ReLM}.

The \textbf{ReaLiSe} model \cite{Xu-etal-2021-realise} employs multi-modal technology to capture semantic, phonetic, and glyph information.
The \textbf{SCOPE} model \cite{Li-etal-2022-SCOPE} is one of the SOTA models for CSC, which enhances model correction performance by introducing a character pronunciation prediction task.
The \textbf{ReLM} model \cite{Liu-etal-2024-ReLM} treats CSC as a non-autoregressive paraphrasing task, standing out as a new SOTA model.

Additionally, we include some of the latest work \cite{Cheng-etal-2020-SpellGCN,Huang-etal-2023-module} for performance comparison.

In the era of LLMs, researchers have begun using LLMs to explore the CSC field.
We present the results of representative LLMs on certain benchmarks for comparison, including the top-performing GPT series in terms of overall capability: \texttt{GPT3.5} and \texttt{GPT4}, as well as some results on open-source LLMs in the Chinese NLP community from previous work, such as finetuned \texttt{Baichuan2} \cite{yang-etal-2023-baichuan2}. 
Specifically, we demonstrate the results of \texttt{Baichuan2} on the ECSpell and LEMON test sets using supervised fine-tuning (SFT) \cite{Liu-etal-2024-ReLM} and prompt-free training-free approach \cite{zhou-etal-2024-llm-csc}, representing the current SOTA performance.

\begin{table}[tb!]
    \setlength{\tabcolsep}{4.5pt}
    \renewcommand{\arraystretch}{0.95}
    \centering
    \fontsize{9}{12}\selectfont
    \begin{NiceTabular}{llcccc}
        \toprule
        \rowcolor[gray]{1.0}
        \Block{2-1}{\textbf{Domain}} & \Block{2-1}{\textbf{Model}} & \Block{1-3}{\textbf{Correction}} & & & \Block{2-1}{$\mathrm{FPR}$} \\ 
        \cmidrule(lr){3-5}
        \rowcolor[gray]{1.0}
        & & $\mathrm{P}$ & $\mathrm{R}$ & $\mathrm{F}_1$ & \\ \midrule
        \rowcolor[gray]{.95}
        \Block{1-6}{ECSpell} & & & & & \\
        \Block{5-1}{LAW} & GPT3.5 & 48.5 & 43.1 & 45.6 & 9.4 \\
        & GPT4 & 62.0 & 62.0 & 62.0 & 7.3 \\
        & Baichuan2-7B$^\star$ & 85.1 & 87.1 & 86.0 & -- \\
        & ReLM & 93.7 & \textbf{98.8} & 96.2 & 6.5 \\
        & \quad + DISC & \textbf{96.5} & 98.0 & \textbf{97.3} & \textbf{2.9} \\ \midrule
        \Block{5-1}{MED} & GPT3.5 & 36.5 & 42.0 & 39.1 & 20.1 \\
        & GPT4 & 45.1 & 57.5 & 50.6 & 24.8 \\
        & Baichuan2-7B$^\star$ & 72.6 & 73.9 & 73.2 & -- \\
        & ReLM & 85.1 & 95.8 & 90.2 & 9.8 \\
        & \quad + DISC & \textbf{91.6} & \textbf{96.3} & \textbf{93.9} & \textbf{4.6} \\ \midrule
        \Block{5-1}{ODW} & GPT3.5 & 57.3 & 52.3 & 54.7 & 6.3 \\
        & GPT4 & 71.7 & 67.6 & 69.5 & \textbf{1.7} \\
        & Baichuan2-7B$^\star$ & 86.1 & 79.3 & 82.6 & -- \\
        & ReLM & 89.4 & \textbf{91.5} & 90.4 & 5.8 \\
        & \quad + DISC & \textbf{91.1} & 91.1 & \textbf{91.1} & 3.3 \\
        \midrule
        \rowcolor[gray]{.95}
        \Block{1-6}{LEMON} & & & & & \\
        \Block{3-1}{GAM} 
        & Baichuan2-7B$^\dagger$ & 38.2 & \textbf{35.6} & 36.9 & 19.8 \\
        & ReLM & 35.8 & 33.6 & 34.6 & 20.6 \\
        & \quad + DISC & \textbf{56.1} & 31.5 & \textbf{40.4} & \textbf{8.5} \\ \midrule
        \Block{3-1}{CAR}
        & Baichuan2-7B$^\dagger$ & 64.3 & 46.8 & 54.2 & 6.9 \\
        & ReLM & 59.2 & \textbf{48.9} & 53.6 & 12.0 \\
        & \quad + DISC & \textbf{72.3} & 45.9 & \textbf{56.2} & \textbf{4.6} \\ \midrule
        \Block{3-1}{ENC} 
        & Baichuan2-7B$^\dagger$ & 59.8 & \textbf{45.1} & \textbf{51.4} & 10.2 \\
        & ReLM & 55.8 & 41.6 & 47.7 & 12.7 \\
        & \quad + DISC & \textbf{72.2} & 39.3 & 50.9 & \textbf{5.1} \\ \midrule
        \Block{3-1}{MEC} 
        & Baichuan2-7B$^\dagger$ & 77.5 & \textbf{49.7} & \textbf{60.6} & 3.4 \\
        & ReLM & 67.3 & 44.9 & 53.9 & 5.8 \\
        & \quad + DISC & \textbf{82.2} & 44.5 & 57.7 & \textbf{2.2} \\ \midrule
        \Block{3-1}{Avg. (all)} 
        & Baichuan2-7B$^\dagger$ & 62.1 & \textbf{46.8} & 53.2 & 9.9 \\
        & ReLM & 58.1 & 45.1 & 50.6 & 11.7 \\
        & \quad + DISC & \textbf{73.7} & 42.5 & \textbf{53.7} & \textbf{4.5} \\
        \bottomrule
    \end{NiceTabular}
    \caption{Sentence-level performance of LLMs, ReLM, and
        ReLM + DISC on the test sets of ECSpell and LEMON. Results marked with ``$\star$'' are from \citet{Liu-etal-2024-ReLM}, and ``$\dagger$'' are from \citet{zhou-etal-2024-llm-csc}.}
    \label{tab:ecspell_experiment}
\end{table}
\subsection{Evaluation Metrics}
The CSC task comprises two subtasks: error detection and error correction.
Following the previous work \cite{zhang-etal-2020-softmasked}, we report the precision ($\mathrm{P}$), recall ($\mathrm{R}$), and $\mathrm{F}_{1}$ scores at the sentence level for both subtasks.
Additionally, we also evaluate the models with the False Positive Rate ($\mathrm{FPR}$) metric \cite{Liu-etal-2024-ReLM}, which quantifies the CSC model's frequency of over-correction, i.e., incorrectly identifying correct sentences as erroneous.

\subsection{Hyperparameters}

Hyperparameters $\alpha$ and $\beta$ denote the weights assigned to overall similarity and phonetic similarity, respectively. As detailed in Section \ref{sec:RobustnessofHyperparameters} on grid search results, we set $\alpha=1.1$ in Eq. \ref{eq:New Distribution} and $\beta=0.7$ in Eq. \ref{eq:Similarity} for all experiments. 

\section{Main Results}
\paragraph{Results on SIGHANs.}
Table \ref{tab:sighans_experiment} illustrates the main results across SIGHAN benchmarks, demonstrating that the addition of the DISC module in the decoding process leads to notable improvements across all the compared models, reaching state-of-the-art performance.
Specifically, ReaLiSe + DISC has increases of 0.1/1.1/0.6, SCOPE + DISC achieves lifts of 1.2/1.4/0.8, and ReLM + DISC sees enhancements of 2.1/3.3/1.2 in correction-level $\mathrm{F}_{1}$ (C-$\mathrm{F}$) score on the SIGHAN13/14/15 test sets, respectively.

It is worth noting that ReaLiSe and SCOPE have incorporated phonetic or glyph information during training.
However, our DISC module can still improve the performance of these models.

In addition to the consistent improvement in the $\mathrm{F}_{1}$ metric, results demonstrate that the integration of the DISC module into CSC models leads to a significant reduction in $\mathrm{FPR}$ across almost all datasets.
This implies that DISC can avoid some unnecessary corrections.

\paragraph{Results on Native Datasets.}
As ReLM has shown outstanding performance on the SIGHAN benchmarks, we continue to utilize it for experiments on the multi-domain datasets of ECSpell and LEMON to demonstrate the DISC module's domain adaptability.

Table \ref{tab:ecspell_experiment} depicts that the incorporation of the DISC module into ReLM leads to substantial improvements of 1.1/3.7/0.7 C-$\mathrm{F}$ score compared to unenhanced ReLM in the LAW, MED and ODW domains, respectively.

Table \ref{tab:ecspell_experiment} also presents the performance of DISC on LEMON.
After integrating the DISC module, the results of ReLM + DISC achieve notable improvements across all domains, and the average C-$\mathrm{F}$ has an increase of 3.1.
This demonstrates that our DISC module yields stable and significant improvements in cross-domain CSC testing.

\begin{figure}[tb!]
    \captionsetup[subfigure]{skip=2pt}
    \begin{center}
        
        \subfloat[Select the more similar word]{
            \centering
            \scalebox{0.96}{
                \begin{tikzpicture}[
                        font=\scriptsize,
                        anchor point/.style={
                                draw=none,
                                circle,
                                inner sep=1.5pt,
                            },
                        word/.style={
                                font=\footnotesize,
                                anchor=base,
                            },
                        bg/.style={
                                fill opacity=0.6,
                                rounded corners=3pt,
                                inner sep=1.2pt,
                                minimum width=0.98\linewidth,
                                minimum height=6 0.0,
                            },
                    ]

                    \node[anchor point] (a west) at (-0.48\linewidth, -2.5) {};
                    \node[anchor point] (a east) at (0.48\linewidth, -2.5) {};
                    \node[anchor point] (a center) at (0, -2.5) {};
                    \node[word, inner sep=0pt, anchor=north west] (source title) at (a west) {\textbf{\texttt{Input:}}};
                    \node[word, anchor=base west] (source) at ($(source title.base east) + (0.6, 0)$) {肌肉酸痛是运动过\textcolor{figure_red}{读(dú)}导致的。};
                    \node[word, anchor=north west, font=\scriptsize, align=left] (source gross) at ($(source.south west) + (0.0, 0.2)$) {\quad \textit{Muscle soreness is caused by \textcolor{figure_red}{read} and exercise. }};
                    \node[word, inner sep=0pt, anchor=north west] (reference title) at ($(source title.south west) + (0, -0.45)$) {\textbf{\texttt{Reference:}}};
                    \node[word, anchor=base west] (reference) at ($(reference title.base-|source.west) + (0, -0.0)$) {\textcolor{figure_red}{读} \ding{212} \textcolor{figure_blue}{度} \scriptsize{(\textit{\textcolor{figure_red}{dú}} \ding{212} \textit{\textcolor{figure_blue}{dù, excessive}})}};
                    \node[word, inner sep=0pt, anchor=north west] (predict title) at ($(reference title.south west) + (0, -0.2)$) {\textbf{\texttt{ReLM:}}};
                    \node[word, anchor=base west] (predict) at ($(predict title.base-|source.west) + (0, -0.0)$) {\textcolor{figure_red}{读} \ding{212} \textcolor{figure_red}{少} \scriptsize{(\textit{\textcolor{figure_red}{dú}} \ding{212} \textit{\textcolor{figure_red}{shǎo, insufficient}})}};
                    \node[word, inner sep=0pt, anchor=north west] (predict title gpt4) at ($(predict title.south west) + (0, -0.2)$) {\textbf{\texttt{ReLM+DISC:}}};
                    \node[word, anchor=base west] (predict gpt4) at ($(predict title gpt4.base-|source.west) + (0, -0.0)$) {\textcolor{figure_red}{读} \ding{212} \textcolor{figure_blue}{度} \scriptsize{(\textit{\textcolor{figure_red}{dú}} \ding{212} \textit{\textcolor{figure_blue}{dù, excessive}})}};
                    \begin{scope}[on background layer]
                        \node[fill=lightgray!30, bg, anchor=north] at ($(a center) + (0.0, 0.15)$) {};
                    \end{scope}
                \end{tikzpicture}
            }
            \label{fig:examples:incorrect}
        }\\[3pt]%
        \subfloat[Mitigate over-correction]{
            \centering
            \scalebox{0.96}{
                \begin{tikzpicture}[
                        font=\scriptsize,
                        anchor point/.style={
                                draw=none,
                                circle,
                                inner sep=1.5pt,
                            },
                        word/.style={
                                font=\footnotesize,
                                anchor=base,
                            },
                        bg/.style={
                                fill opacity=0.6,
                                rounded corners=3pt,
                                inner sep=1.2pt,
                                minimum width=0.98\linewidth,
                                minimum height=60.0,
                            },
                    ]
                    \node[anchor point] (a west) at (-0.48\linewidth, 0) {};
                    \node[anchor point] (a east) at (0.48\linewidth, 0) {};
                    \node[anchor point] (a center) at (0, 0) {};
                    \node[word, inner sep=0pt, anchor=north west] (source title) at (a west) {\textbf{\texttt{Input:}}};
                    \node[word, anchor=base west] (source) at ($(source title.base east) + (0.6, 0)$) {浓荫蔽\textcolor{figure_blue}{空(kōng)}，郁郁苍苍。};
                    \node[word, anchor=north west, font=\scriptsize, align=left] (source gross) at ($(source.south west) + (0.0, 0.2)$) {\quad \textit{Thick foliage shades the \textcolor{figure_blue}{sky}, lush and verdant.}};
                    \node[word, inner sep=0pt, anchor=north west] (reference title) at ($(source title.south west) + (0, -0.45)$) {\textbf{\texttt{Reference:}}};
                    \node[word, anchor=base west] (reference) at ($(reference title.base-|source.west) + (0, -0)$) {\texttt{NONE}};
                    \node[word, inner sep=0pt, anchor=north west] (predict title) at ($(reference title.south west) + (0, -0.2)$) {\textbf{\texttt{ReLM:}}};
                    \node[word, anchor=base west] (predict) at ($(predict title.base-|source.west) + (0, -0.05)$) {\textcolor{figure_blue}{空} \ding{212} \textcolor{figure_red}{日}\scriptsize{(\textit{\textcolor{figure_blue}{kōng}} \ding{212} \textit{\textcolor{figure_red}{rì, sun}})}};
                    \node[word, inner sep=0pt, anchor=north west] (predict title gpt4) at ($(predict title.south west) + (0, -0.2)$) {\textbf{\texttt{ReLM+DISC:}}};
                    \node[word, anchor=base west] (predict gpt4) at ($(predict title gpt4.base-|source.west) + (0, 0)$) {\texttt{NONE}};

                    \begin{scope}[on background layer]
                        \node[fill=lightgray!30, bg, anchor=north] at ($(a center) + (0.0, 0.15)$) {};
                    \end{scope}
                \end{tikzpicture}
            }
            \label{fig:examples:correct}
        }
    \end{center}
    \caption{Cases from the SIGHANs and LEMON.}
    \label{fig:intro:examples}
\end{figure}
\subsection{Case Study}
\label{sec:appendix:case_study}

We present two illustrative examples of DISC-augmented error correction in Figure \ref{fig:intro:examples}.
These examples explain why our DISC module can significantly improve model precision.

Figure \ref{fig:examples:incorrect} exemplifies how the DISC module retrieves a more plausible alteration resembling the original character.
In this example, the ReLM model corrects the erroneous word ``读''(dú) to ``少''(shǎo).
This correction is grammatically correct, but deviates from the original meaning of the sentence.
From the perspective of phonetics, a more suitable correction should be ``度''(dù), which shares the same pronunciation as the erroneous word.
The DISC makes this correction by leveraging semantic and phonetic information.

In Figure \ref{fig:examples:correct}, the DISC alleviates over-correction. The CSC model mistakenly alters ``空''(kōng) to ``日''(rì), yet the similarity intervention rectifies this error.
Specifically, since the most similar to a character is the character itself, when a CSC model incorrectly tends to correct over preserve on a correct sentence, the DISC module can increase the score of the character itself compared to other correction options based on similarity, which sometimes avoids unnecessary corrections.

\section{Discussion}
We select the SIGHAN15 along with two domains from the LEMON database, ENC and MEC, to conduct further analysis.

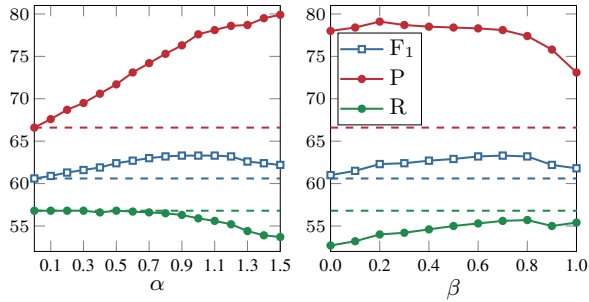
\begin{figure}[tb!]
    \centering
    \hspace*{1.7mm}
    \scalebox{0.95}{
    \begin{tikzpicture}[
            legend/.style={
                    fill=white,
                    font=\footnotesize,
                    inner sep=2pt,
                    minimum width=0.8cm,
                    text opacity=1.0,
                    fill opacity=1.0,
                },
            diff label/.style={
                    font=\scriptsize,
                    inner sep=0.5pt,
                    outer sep=1.5pt,
                    fill=white,
                    fill opacity=0.9,
                    text opacity=1.0,
                    rounded corners=1pt,
                },
            trim left
        ]
        \centering
        \begin{groupplot}[
                group style={
                        group size=2 by 1,
                        x descriptions at=edge bottom,
                        horizontal sep=0.7cm,
                        vertical sep=0.1cm,
                    },
                width=0.65\linewidth,
                height=0.65\linewidth,
                xlabel={$\alpha$},
                xmin=0.0,
                xmax=1.5,
                xtick={0.1, 0.3, 0.5, 0.7, 0.9, 1.1, 1.3, 1.5},
                xticklabels={0.1, 0.3, 0.5, 0.7, 0.9, 1.1, 1.3, 1.5},
                ytick={55, 60, 65, 70, 75, 80},
                /tikz/font=\scriptsize,
                ylabel shift=-4pt,
                xlabel shift=-4pt,
                xlabel style={font=\footnotesize}, 
                ylabel style={font=\footnotesize}, 
                yticklabel shift=-2pt,
                xticklabel shift=-1pt,
                legend style={
                        at={(0.98,0.45)},
                        anchor=east,
                        font=\footnotesize
                    },
            ]
            \nextgroupplot[ymin=52,ymax=81]
            \addplot+ [mark=square*, draw=figure_blue, thick,
                mark size=1.25pt,
                mark options={fill=white, fill opacity=1.0, solid},
                opacity=1.0,
            ] table [row sep=\\] {
                    x	y\\
                    0.0	60.6\\
                    0.1	60.9\\
                    0.2	61.3\\
                    0.3	61.6\\
                    0.4	61.9\\
                    0.5	62.4\\
                    0.6	62.7\\
                    0.7	63.0\\
                    0.8	63.2\\
                    0.9	63.3\\
                    1.0	63.3\\
                    1.1	63.3\\
                    1.2	63.2\\
                    1.3	62.6\\
                    1.4	62.4\\
                    1.5	62.2\\
                };
            \addplot+ [mark=*, draw=figure_red, thick,
                mark size=1.25pt,
                mark options={fill=figure_red, fill opacity=1.0, solid},
                opacity=1.0,
            ] table [row sep=\\] {
                    x	y\\
                    0.0	66.6\\
                    0.1	67.6\\
                    0.2	68.7\\
                    0.3	69.5\\
                    0.4	70.6\\
                    0.5	71.7\\
                    0.6	73.1\\
                    0.7	74.2\\
                    0.8	75.3\\
                    0.9	76.3\\
                    1.0	77.6\\
                    1.1	78.1\\
                    1.2	78.6\\
                    1.3	78.7\\
                    1.4	79.5\\
                    1.5	79.9\\
                };
            \addplot+ [mark=*, draw=figure_green, thick,
                mark size=1.25pt,
                mark options={fill=figure_green, fill opacity=1.0, solid},
                opacity=1.0,
            ] table [row sep=\\] {
                    x	y\\
                    0.0	56.8\\
                    0.1	56.8\\
                    0.2	56.8\\
                    0.3	56.8\\
                    0.4	56.6\\
                    0.5	56.8\\
                    0.6	56.7\\
                    0.7	56.6\\
                    0.8	56.5\\
                    0.9	56.3\\
                    1.0	55.9\\
                    1.1	55.6\\
                    1.2	55.2\\
                    1.3	54.4\\
                    1.4	53.9\\
                    1.5	53.7\\
                };
            \draw[draw=figure_blue, thick, dashed] (axis cs:0.0,60.6) -- (axis cs:1.5,60.6);
            \draw[draw=figure_red, thick, dashed] (axis cs:0.0,66.6) -- (axis cs:1.5,66.6);
            \draw[draw=figure_green, thick, dashed] (axis cs:0.0,56.8) -- (axis cs:1.5,56.8);
            \nextgroupplot[ymin=52,ymax=81, xmin=0.0,
                xmax=1.0,
                xlabel={$\beta$},
                xtick={0.0, 0.2, 0.4, 0.6, 0.8, 1.0},
                xticklabels={0.0, 0.2, 0.4, 0.6, 0.8, 1.0},
                legend style={
                        at={(0.2,0.89)}, 
                        anchor=north, 
                        font=\footnotesize,
                        inner sep=0.5pt,
                        minimum width=0.1cm,
                        text opacity=1.0,
                        fill opacity=1.0,
                        cells={anchor=west}, 
                        scale=0.01 
                    },
                ]
            \addplot+ [mark=square*, draw=figure_blue, thick,
                mark size=1.25pt,
                mark options={fill=white, fill opacity=1.0, solid},
                opacity=1.0,
            ] table [row sep=\\] {
                    x	y	ey+	ey-\\
                    0.0	61.0\\
                    0.1	61.5\\
                    0.2	62.3\\
                    0.3	62.4\\
                    0.4	62.7\\
                    0.5	62.9\\
                    0.6	63.2\\
                    0.7	63.3\\
                    0.8	63.2\\
                    0.9	62.2\\
                    1.0	61.8\\
                };
            \addplot+ [mark=*, draw=figure_red, thick,
                mark size=1.25pt,
                mark options={fill=figure_red, fill opacity=1.0, solid},
                opacity=1.0,
            ] table [row sep=\\] {
                    x	y\\
                    0.0	78.0\\
                    0.1	78.4\\
                    0.2	79.1\\
                    0.3	78.7\\
                    0.4	78.5\\
                    0.5	78.4\\
                    0.6	78.3\\
                    0.7	78.1\\
                    0.8	77.4\\
                    0.9	75.8\\
                    1.0	73.1\\
                };
            \addplot+ [mark=*, draw=figure_green, thick,
                mark size=1.25pt,
                mark options={fill=figure_green, fill opacity=1.0, solid},
                opacity=1.0,
            ] table [row sep=\\] {
                    x	y\\
                    0.0	52.7\\
                    0.1	53.2\\
                    0.2	54.0\\
                    0.3	54.2\\
                    0.4	54.6\\
                    0.5	55.0\\
                    0.6	55.3\\
                    0.7	55.6\\
                    0.8	55.7\\
                    0.9	55.0\\
                    1.0	55.4\\
                };
            \legend{$\mathrm{F}_1$, $\mathrm{P}$, $\mathrm{R}$};
            \draw[draw=figure_blue, thick, dashed] (axis cs:0.0,60.6) -- (axis cs:1.0,60.6);
            \draw[draw=figure_red, thick, dashed] (axis cs:0.0,66.6) -- (axis cs:1.5,66.6);
            \draw[draw=figure_green, thick, dashed] (axis cs:0.0,56.8) -- (axis cs:1.5,56.8);
        \end{groupplot}
    \end{tikzpicture}
    }
    \caption{The average scores in ENC, MEC and SIGHAN15 with different values of $\alpha$ and $\beta$. The solid lines represent the results of ReLM + DISC, and the dashed lines represent the results of the original ReLM. }
    \label{fig:alpha_and_beta}
\end{figure}

\paragraph{Robustness of similarity hyperparameters.}
\label{sec:RobustnessofHyperparameters}
As illustrated in Figure \ref{fig:alpha_and_beta}, the model's precision steadily improves as $\alpha$ increases. This is because increased similarity intervention reduces over-correction (Figure \ref{fig:examples:incorrect}), boosting precision.
However, at the same time, DISC may revert predictions to the original character, as characters are most similar to themselves. This under-correction phenomenon caused by DISC sometimes leads to instability in recall. 

For $\beta$, it shows the effect of the proportion of phonetic similarity in the total similarity on correction performance. The $\mathrm{F}_1$ score curve shows a clear trend of rising first and then decreasing, which indicates that phonetic and glyph similarities are complementary, with phonetic similarity being relatively more important than glyph similarity.

\paragraph{The balance between precision and recall.}
The primary purpose of using a confusion set is to narrow down the retrieval space, thereby improving precision.
While a confusion set can enhance recall by enabling the model to make more reasonable edits, it may also reduce recall by discouraging the model from making edits, because the most similar character to the source character is itself.

We observe that this decrease in recall primarily occurs in the LEMON test set.
The key distinction between these datasets is that LEMON contains less seen edit pairs in training data.
As shown in Table~\ref{tab:proportion_of_seen_pairs}, we calculate the proportion of edit pairs from each test set that appear in the training set.\footnote{For LEMON, we conduct statistics using the pairs in the confusion set which is used to generate 34 million monolingual sentences.}
In ECSpell, the proportion of seen edit pairs is very low, yet the model performs well. This is due to data leakage, as we explain in Appendix~\ref{sec:appendix:cleaned_ecspell}.
Models are prone to copy the source character when the target edit pair is not available in the training data.

For this type of test set, we can use a simple copy punishment combined with DISC, which reduces the probability of copying the original character during inference, to mitigate the decrease in recall.
Detailed experimental results can be found in Appendix~\ref{sec:appendix:recall}.
\begin{table}[tb!]
    \centering
    \renewcommand{\arraystretch}{0.95}
    \fontsize{9}{12}\selectfont
    \begin{tabular}{cccc}
        \toprule[1pt]
        \textbf{Domain} & \textbf{Edit Pairs} & \textbf{Seen Pairs} & \textbf{Prop.} \\ \midrule
        \multicolumn{4}{c}{\cellcolor[HTML]{F2F2F2}SIGHANs} \\
        SIGHAN15 & \wz703 & \wz698 & 99.29\% \\
        SIGHAN14 & \wz771 & \wz765 & 99.22\% \\
        SIGHAN13 & 1,224 & 1,206 & 98.53\% \\ \midrule
        \multicolumn{4}{c}{\cellcolor[HTML]{F2F2F2}ECSpell}  \\
        LAW      & 390 & 211 & 54.10\% \\
        MED      & 356 & 169 & 47.47\% \\
        ODW      & 404 & 168 & 41.58\% \\ \midrule
        \multicolumn{4}{c}{\cellcolor[HTML]{F2F2F2}LEMON}  \\
        GAM      & 164  & 100  & 60.98\% \\
        CAR      & 1,911 & 1,254 & 65.62\% \\
        NOV      & 3,415 & 2,045 & 59.88\% \\
        ENC      & 1,787 & 1,040 & 58.20\% \\
        NEW      & 3,260 & 2,293 & 70.34\% \\
        COT      & 486  & 309  & 63.58\% \\
        MEC      & 1,032 & 627  & 60.76\% \\
        \bottomrule[1pt]
    \end{tabular}
    \caption{Proportion of seen edit pairs in the test sets of SIGHANs, ECSpell, and LEMON. }
    \label{tab:proportion_of_seen_pairs}
\end{table}


\paragraph{Effectiveness of DISC module.}
We conducted experiments using the vanilla BERT without fine-tuning, initialized with \texttt{bert-base-chinese}, as shown in Table~\ref{tab:vanilla_BERT}. Obviously, the general language model, without any fine-tuning or error correction strategies, cannot be applied to error correction tasks. However, after adding our DISC module, the performance of vanilla BERT improves significantly, giving the model basic error correction capabilities.

To further demonstrate the superiority of our method, we degrade the DISC module to a simple confusion set constraint decoding strategy.
We investigate two confusion sets: one derived from our similarity computation strategy\footnote{We treat a character pair as confused if their similarity score exceeds 0.5.} and another pre-existing one provided by \citet{wang-etal-2018-hybrid}.
The results are shown in the second part of Table~\ref{tab:ablation_lemon}.
From the results, we can see that both confusion sets fail to consistently improve performance, indicating the strategy’s sensitivity to confusion set quality. 
The confusion set from \citet{wang-etal-2018-hybrid} improves SIGHAN15 by covering over 99\% of its erroneous pairs but degrades performance on other test sets, highlighting the domain-specific limitations of such confusion sets.

\paragraph{Effectiveness of components of the DISC module.}
We conduct an ablation study on the components of the DISC module.
The results are shown in the third part of Table \ref{tab:ablation_lemon}.
Removing either phonetic or glyph knowledge from the DISC module leads to performance declines across benchmarks.
Notably, the absence of phonetic similarity has a lesser effect on SIGHAN15 but a stronger impact on LEMON.
The results also show that the four components involved in calculating glyph similarity are independently effective.
However, excluding any three typically causes a slight drop in performance, with exceptions like ENC.
This phenomenon underscores the necessity of using multi-dimensional similarity measurements for a more comprehensive modeling of glyph similarity.
Combining these often results in consistent improvements. 
Moreover, the fusion of phonetic and glyph similarities achieves the optimal error correction performance, affirming the necessity of integrating these two similarities.

\paragraph{Analysis on Different Error Types.}
Based on the similarity calculation strategy proposed in this paper, we separately filter the phonetic and glyph error types. Taking the phonetic error type as an example, the specific approach is as follows: for all edit pairs with a phonetic similarity < 0.5, we modify the source character to the golden character, resulting in a test set containing only phonetic edit pairs. The results are shown in Table~\ref{tab:different_error_types}.

According to our classification rule, the average proportions of phonetic and glyph on SIGHAN15 are 90.0\% and 40.7\%. From the results, it can be seen that DISC significantly improves performance on all error types. We find that, compared to phonetic errors, the original model exhibited a higher Recall metric and lower Precision metric on glyph errors, which makes the improvement brought by the DISC module more pronounced.

\begin{table}[tb!]
    \setlength{\tabcolsep}{4.5pt}
    \renewcommand{\arraystretch}{0.95}
    \centering
    \fontsize{9}{12}\selectfont
    \begin{NiceTabular}{lccccc}
        \toprule
        \rowcolor[gray]{1.0}
        \Block{2-1}{\textbf{Domain}} & \Block{1-3}{\textbf{Correction}} & & & \Block{2-1}{$\mathrm{FPR}$} \\ 
        \cmidrule(lr){2-4}
        \rowcolor[gray]{1.0}
        & $\mathrm{P}$ & $\mathrm{R}$ & $\mathrm{F}_1$ & \\ \midrule
        \rowcolor[gray]{.95}
        \Block{1-5}{SIGHAN15} & & & & \\
        vanilla BERT & 2.3 & 4.4 & 3.1 & 91.1 \\
        \quad + DISC & \textbf{71.5} & \textbf{25.5} & \textbf{37.6} & \textbf{2.1} \\ \midrule
        \rowcolor[gray]{.95}
        \Block{1-5}{LEMON-ENC} & & & & \\
        vanilla BERT & 3.6 & 5.8 & 4.4 & 73.0 \\
        \quad + DISC & \textbf{79.9} & \textbf{18.3} & \textbf{29.8} & \textbf{1.0} \\ \midrule
        \rowcolor[gray]{.95}
        \Block{1-5}{LEMON-MEC} & & & & \\
        vanilla BERT & 2.7 & 4.8 & 3.5 & 77.2 \\
        \quad + DISC & \textbf{88.4} & \textbf{18.5} & \textbf{30.5} & \textbf{0.3} \\
        \bottomrule
    \end{NiceTabular}
    \caption{Sentence-level performance of vanilla BERT.}
    \label{tab:vanilla_BERT}
\end{table}
\paragraph{Impact on Decoding Efficiency.}
We examine the influence of the DISC module on decoding speed, with the results shown in Table \ref{tab:decoding_efficiency}.
Phonetic and glyph similarities can be pre-calculated and DISC only need to index them during decoding.
Thus, the time taken to decode each sentence increased merely by 14.3\%, 3.5\%, and 1.0\% for ReaLiSe, SCOPE, and ReLM, respectively. The minor slowdown in decoding speed incurred by the DISC module is deemed acceptable considering the substantial enhancement it brings to the model's performance. Notably, SCOPE exhibits significantly slower decoding speeds compared to the other two models, which we speculate may be attributed to its iterative decoding approach.

\begin{table}[tb!]
\centering
\fontsize{10}{13}\selectfont
\setlength{\tabcolsep}{0.1cm}
\begin{tabular}{lcccc}
\toprule[1pt]
\textbf{Model}              & ENC    & MEC    & SIG15      & Avg         \\ \midrule
ReLM                        & 47.7          & 53.9          & 80.2          & 60.6          \\
\quad + DISC             & 50.9          & \textbf{57.7} & \textbf{81.4} & \textbf{63.3} \\ \midrule
\quad + Confusion set & 47.1          & 56.0          & 78.0          & 60.4          \\
\quad + Confusion set$^\ddagger$        & 41.5          & 48.7          & 80.7          & 57.0          \\ \midrule
\quad + DISC (phonetic)        & 49.1          & 56.1          & 80.1          & 61.8          \\
\quad + DISC (glyph)           & 49.4          & 53.3          & 80.3          & 61.0          \\
\quad + DISC (phonetic \&)        &               &               &               &               \\
\quad\quad├ $\texttt{Sim}^{\texttt{G}}_1$                   & 50.5          & 56.8          & \textbf{81.4}          & 62.9          \\
\quad\quad├ $\texttt{Sim}^{\texttt{G}}_2$                   & 50.5          & 57.4          & \textbf{81.4}          & 63.1          \\
\quad\quad├ $\texttt{Sim}^{\texttt{G}}_3$                   & 51.3          & 57.5          & 81.2          & \textbf{63.3}          \\
\quad\quad└ $\texttt{Sim}^{\texttt{G}}_4$                   & \textbf{51.6} & 56.9          & 80.8          & 63.1          \\ \bottomrule[1pt]
\end{tabular}
\caption{Ablation results in two kinds of confusion sets and different components of DISC. 
``$\ddagger$'' represents the confusion set from \citet{wang-etal-2018-hybrid}.
``$\texttt{Sim}^{\texttt{G}}_i$'' means using similarities of phonetics and the $i$th part of glyph.}
\label{tab:ablation_lemon}
\end{table}
    \section{Related Work}

\paragraph{Model architecture shift.} 
Most early works on CSC employed a three-step pipeline, i.e., 1) detecting potential erroneous characters, 2) constructing new sentences by replacing erroneous characters with new ones based on a confusion set; and 3) evaluating the probability of the constructed sentences based on an $n$-gram language model and choose the one with the highest probability \cite{yeh-etal-2013-chinese,yu-li-2014-chinese,huang-etal-2014-chinese,xie-etal-2015-chinese}.

In the current deep-learning era, especially with the prevalence of PLMs, 
recent models directly perform character-level replacement via classification, as introduced in Section \ref{sec:basic_approach:basic_csc_model}. 
There also exist some works that employ a two-step pipeline architecture, which first detects potentially erroneous characters and then replaces them at the detected positions \citep{zhang-etal-2020-softmasked,Huang-etal-2023-module}. 

\paragraph{Utilizing confusion sets.} These works fall into three categories. 
(1) \emph{At only the inference phase.}
\citet{wang-etal-2019-confusionset} and \citet{Bao-etal-2020-confusionset} use the confusion set as constraints upon the search space, i.e., allowing the model to only consider characters in the confusion set.

(2) \emph{For data synthesis.} \citet{Liu-etal-2021a-pretrain} use a confusion set $\mathcal{C}$ to synthesize data for training CSC models. For a given correct sentence, they randomly select a character (e.g., $c_i$), and replace it with an incorrect character (e.g., $c'$).  
The replacement is constrained such that only pairs contained in the confusion set are considered, i.e., $(c_i, c') \in \mathcal{C}$.

(3) \emph{At both training and inference phases.} \citet{Cheng-etal-2020-SpellGCN} construct two character graphs, one based on phonetic relatedness, and the other based on glyph relatedness, and employ GCN to obtain new character representations as extra inputs. 
\citet{Huang-etal-2023-module} use two confusion sets, one encoding phonetic relatedness, and the other encoding glyph relatedness. Given a potential spelling error, they use a classification module to judge which confusion set the error belongs to, with an extra training loss. During the test phase, the model can only consider characters from the corresponding confusion set according to the classification result. 

\begin{table}[tb!]
    \setlength{\tabcolsep}{4.5pt}
    \renewcommand{\arraystretch}{0.95}
    \centering
    \fontsize{9}{12}\selectfont
    \begin{NiceTabular}{lccccc}
        \toprule
        \rowcolor[gray]{1.0}
        \Block{2-1}{\textbf{Domain}} & \Block{1-3}{\textbf{Correction}} & & & \Block{2-1}{$\mathrm{FPR}$} \\ 
        \cmidrule(lr){2-4}
        \rowcolor[gray]{1.0}
        & $\mathrm{P}$ & $\mathrm{R}$ & $\mathrm{F}_1$ & \\ \midrule
        \rowcolor[gray]{.95}
        \Block{1-5}{Phonetic errors} & & & & \\
        ReLM & 65.2 & 77.6 & 70.9 & 19.3 \\
        \quad + DISC & \textbf{69.8} & \textbf{78.0} & \textbf{73.7} & \textbf{14.3} \\ \midrule
        \rowcolor[gray]{.95}
        \Block{1-5}{Glyph errors} & & & & \\
        ReLM & 44.7 & 81.5 & 57.7 & 19.6 \\
        \quad + DISC & \textbf{51.8} & \textbf{83.4} & \textbf{63.9} & \textbf{15.1} \\
        \bottomrule
    \end{NiceTabular}
    \caption{Sentence-level performance of different error types on SIGHAN15.}
    \label{tab:different_error_types}
\end{table}

\paragraph{Utilizing phonetic and glyph information.}
Besides the use of confusion sets, there exist some works that directly utilize phonetic and glyph information to enhance CSC models. 
\citet{Liu-etal-2021a-pretrain} and \citet{Li-etal-2022-SCOPE} add an extra task of predicting the phonetic of each input character. 
\citet{Xu-etal-2021-realise} use GRU to encode Pinyin, and use CNN to encode glyphs (font pictures) for each input character, as extra character representations.

\paragraph{Decoding intervention.}
\citet{wei-etal-2021-thinktwice} extract features such as probability and rank of the original character and the top 1 candidate character, and use SVM to judge modification retention.
\citet{yin-etal-2024-error} retrieve similar segments from the training set, and intervene in the decoding process based on the segment (n-gram) similarity between the retrieved segments and the input.
\citet{Lv-etal-2023-ECSpell} employ a word dictionary in the target domain to assist the decoding process. 

\begin{table}[t!]
    \centering
    \fontsize{10}{11}\selectfont
    \begin{tabular}{lcc}
    \toprule[1pt]
    Model & Speed (ms/sent) \bigstrut & Slowdown \bigstrut \\ \midrule
    ReaLiSe        & 24.5           & --           \\
    \quad + DISC         & 27.5           & 1.143$\times$         \\
    \hline
    SCOPE          & 138.6          & --           \\
    \quad + DISC         & 143.4           & 1.035$\times$         \\
    \hline
    ReLM           & 12.7             & --           \\
    \quad + DISC         & 12.8           & 1.010$\times$ \\ \bottomrule[1pt]
    \end{tabular}
    \caption{The decoding time per sentence with a batch size of 1 on SIGHAN15. The results are the average time of three runs.}
    \label{tab:decoding_efficiency}
\end{table}
    \section{Conclusions}
We propose a plug-and-play decoding intervention strategy that enhances CSC models by utilizing phonetic and glyph similarities through a tailored algorithm. Unlike methods that alter model training, our training-free strategy only modifies the decoding process, making it adaptable to almost all mainstream CSC models. Experiments on multiple CSC benchmarks demonstrate that our method significantly improves baselines, and even surpasses the current SOTA models. 
Furthermore, experimental analyses demonstrate that our DISC module helps the model better identify similar candidate characters, effectively reducing over-correction.
Our research has transcended the limitations of traditional confusion set decoding intervention, proving that specific measures and combinations of phonetic and glyph similarities are necessary.

\section*{Limitations}
We believe that our work can be further improved from two aspects. 
First, our experiments focus on the CSC datasets, while our approach can apply to other languages such as Japanese and Korean. 
Second, as a general-use technique, our proposed approach for determining character similarity may not be optimal for CSC in specific domains or scenarios. In that case, we may need to consider more factors besides phonetic and glyph information to compute character similarity. 

\section*{Acknowledgements}
We are deeply grateful to the anonymous reviewers for their thoughtful comments and dedicated efforts, which have greatly contributed to enhancing the quality and clarity of our work.

This work was supported by National Natural Science Foundation of China (Grant No. 62036004 and 62176173), Alibaba Group through Alibaba Innovative Research Program, and Project Funded by the Priority Academic Program Development of Jiangsu Higher Education Institutions.

    \bibliography{custom}

    \appendix
    \appendix
\section{Implementation Details}
\label{sec:appendix:details}
We use the official implementation of ReaLiSe and directly utilize the checkpoint provided by its GitHub repository,\footnote{\url{https://github.com/DaDaMrX/ReaLiSe}}
which initializes the semantic encoder
with the weights of \texttt{chinese- roberta-wwm-ext}.\footnote{\url{https://huggingface.co/hfl/chinese-roberta-wwm-ext}} ReLM uses the official BERT weights \texttt{bert-base-chinese},\footnote{\url{https://huggingface.co/bert-base-chinese}} and only offered the checkpoint after pre-training in 34 million monolingual sentences that are synthesized by confusion set. We fine-tune it on SIGHANs and ECSpell with a batch size of 128 and a learning rate of 3e-5, and the MFT strategy \cite{Wu-etal-2023c-LEMON} is used during training. SCOPE utilizes the pre-trained weights from the \texttt{ChineseBERT-base},\footnote{\url{https://huggingface.co/ShannonAI/ChineseBERT-base}} and we leverage their official implementation for fine-tuning.\footnote{\url{https://github.com/jiahaozhenbang/SCOPE}}
We did not attempt DR-CSC + DISC because they have not fully open-sourced their work. Due to our decoding intervention strategy being deterministic, without any random factors, the experiments are conducted only once. All experiments are conducted on one Tesla V100S-PCIE-32GB GPU.


\section{Details of Datasets}

\label{sec:appendix:datasets}
\paragraph{SIGHANs.}
Following the setup of previous work, we employ SIGHAN 13/14/15 datasets \cite{wu-etal-2013-chinese,yu-etal-2014-overview,tseng-etal-2015-introduction} as our training sets, in conjunction with Wang271K \cite{wang-etal-2018-hybrid}, which consists of 271K synthetically generated instances. We employ the test sets of SIGHAN13/14/15 for evaluation.
\paragraph{ECSpell.}
ECSpell \cite{Lv-etal-2023-ECSpell} encompasses data from three domains: law, medical treatment, and official document writing. Unlike SIGHANs from Chinese learner texts, the sentences in ECSpell are derived from CNS texts.
\paragraph{LEMON.}
LEMON \cite{Wu-etal-2023c-LEMON} also originates from CNS texts, containing over 22K instances spanning 7 domains. Given its lack of a dedicated training set, LEMON serves as a benchmark for evaluating the domain adaptation capability of CSC models.

We conduct detailed statistics on the above datasets, and the results are presented in Table \ref{tab:datasets_details}.
\begin{table}[tb!]
\centering
\resizebox{0.48\textwidth}{!}{%
\begin{tabular}{lrrr}
\toprule[1pt]
Training Set & \#Sent  & Avg. Length & \#Errors \bigstrut  \\ \midrule
SIGHAN15              & 2,339   & 31.3        & 2,549    \\
SIGHAN14              & 3,437   & 49.6        & 3,799    \\
SIGHAN13              & 700     & 41.8        & 343      \\
Wang271K              & 271,329 & 42.6        & 381,962  \\
ECSpell\_LAW          & 1,960   & 30.7        & 1,681    \\
ECSpell\_MED          & 3,000   & 50.2        & 2,260    \\
ECSpell\_ODW          & 1,720   & 41.2        & 1,578    \\ \midrule\midrule
Test Set              & \#Sent  & Avg. Length & \#Errors 
 \bigstrut \\ \midrule
SIGHAN15              & 1,100   & 30.6        & 703      \\
SIGHAN14              & 1,062   & 50.0        & 771      \\
SIGHAN13              & 1,000   & 74.3        & 1,224    \\
ECSpell\_LAW          & 500     & 29.7        & 390      \\
ECSpell\_MED          & 500     & 49.6        & 356      \\
ECSpell\_ODW          & 500     & 40.5        & 404      \\
LEMON          & 22,252     & 35.4        & 12,055      \\\bottomrule[1pt]
\end{tabular}}
\caption{Statistics of the datasets.}
\label{tab:datasets_details}
\end{table}

\section{Copy Punishment}
\label{sec:appendix:recall}
For datasets like LEMON that lack in-domain training data, we discover a simple recall-boosting solution: reducing the probability of selecting the original character during inference. Specifically, after incorporating the DISC module, we additionally lower the prediction probability of the original character by 0.1 to reduce the model's tendency to select the original character during inference, thereby improving the model's recall rate. The experimental results can be found in Table~\ref{tab:recall_solution}.

\begin{table}[tb!]
    \centering
    \fontsize{9}{12}\selectfont
    \begin{NiceTabular}{llccccc}
        \toprule
        \rowcolor[gray]{1.0}
        \Block{2-1}{\textbf{Domain}} & \Block{2-1}{\textbf{Model}} & \Block{1-3}{\textbf{Correction}} & & & \Block{2-1}{$\mathrm{FPR}$} \\ 
        \cmidrule(lr){3-5}
        \rowcolor[gray]{1.0}
        & & $\mathrm{P}$ & $\mathrm{R}$ & $\mathrm{F}_1$ & \\ 
        \midrule
        \rowcolor[gray]{.95}
        \Block{1-6}{LEMON} & & & & & \\
        \Block{3-1}{GAM} & ReLM & 35.8 & 33.6 & 34.6 & 20.6 \\
        & \quad + DISC & \textbf{56.1} & 31.5 & 40.4 & \textbf{8.5} \\
        & \quad + DISC\rlap{*} & 52.4 & \textbf{37.0} & \textbf{43.4} & 11.3 \\ \midrule
        \Block{3-1}{CAR} & ReLM & 59.2 & \textbf{48.9} & 53.6 & 12.0 \\
        & \quad + DISC & \textbf{72.3} & 45.9 & \textbf{56.2} & \textbf{4.6} \\
        & \quad + DISC\rlap{*} & 68.0 & 47.5 & 55.9 & 6.2 \\ \midrule
        \Block{3-1}{NOV} & ReLM & 46.3 & \textbf{32.2} & 38.0 & 17.6 \\
        & \quad + DISC & \textbf{65.2} & 29.6 & \textbf{40.8} & \textbf{7.1} \\
        & \quad + DISC\rlap{*} & 57.8 & 31.2 & 40.6 & 10.2 \\ \midrule
        \Block{3-1}{ENC} & ReLM & 55.8 & 41.6 & 47.7 & 12.7 \\
        & \quad + DISC & \textbf{72.2} & 39.3 & 50.9 & \textbf{5.1} \\
        & \quad + DISC\rlap{*} & 66.9 & \textbf{41.7} & \textbf{51.4} & 7.1 \\ \midrule
        \Block{3-1}{NEW} & ReLM & 68.5 & \textbf{51.5} & 58.8 & 8.4 \\
        & \quad + DISC & \textbf{80.4} & 48.1 & 60.2 & \textbf{3.2} \\
        & \quad + DISC\rlap{*} & 76.5 & 49.7 & \textbf{60.3} & 4.3 \\ \midrule
        \Block{3-1}{COT} & ReLM & 73.5 & \textbf{62.8} & 67.7 & 4.9 \\
        & \quad + DISC & \textbf{87.4} & 58.3 & \textbf{69.9} & \textbf{1.1} \\
        & \quad + DISC\rlap{*} & 80.8 & 61.0 & 69.5 & 1.8 \\ \midrule
        \Block{3-1}{MEC} & ReLM & 67.3 & 44.9 & 53.9 & 5.8 \\
        & \quad + DISC & \textbf{82.2} & 44.5 & \textbf{57.7} & \textbf{2.2} \\
        & \quad + DISC\rlap{*} & 76.3 & \textbf{45.0} & 56.6 & 3.2 \\ 
        \bottomrule
    \end{NiceTabular}
    \caption{Sentence-level performance of LLMs, ReLM, and
        ReLM + DISC on the test sets of ECSpell and LEMON. Results marked with ``*'' indicate the use of copy-punishment solution.}
    \label{tab:recall_solution}
\end{table}


\section{Prompt Example}
\label{sec:appendix:Prompt Examples}
\begin{figure}[tb]
    \centering%
    \newtcolorbox{promptbox}[1]{%
        left=0pt,
        right=0pt,
        top=0pt,
        bottom=0pt,
        boxsep=6pt,
        colframe=black,
        title={#1},
    }
    \begin{minipage}[b]{\columnwidth}
        \begin{promptbox}{\textbf{System and User Prompts for LLMs}}
            \large
            \scriptsize
            \textbf{\texttt{System Prompt:}}\\
            你是一个优秀的中文拼写纠错模型，中文拼写纠错模型即更正用户输入句子中的拼写错误。
            
            \textbf{\texttt{User Prompt:}}\\
            你需要识别并纠正用户输入的句子中可能的错别字并输出正确的句子，纠正时必须保证改动前后句子等长，在纠正错别字的同时尽可能减少对原句子的改动(不添加额外标点符号，不添加额外的字，不删除多余的字)。只输出没有错别字的句子，不要添加任何其他解释或说明。如果句子没有错别字，就直接输出和输入相同的句子。
        \end{promptbox}
    \end{minipage}
    \caption{Prompt template used in GPT3.5 and GPT4.}
    \label{fig:prompt_examples}
\end{figure}

In this work, we use the prompt-based method to activate the CSC ability of the \texttt{GPT3.5} and \texttt{GPT4}. The prompt is shown in Figure \ref{fig:prompt_examples}.

\section{Experiments on Cleaned ECSpell}
\label{sec:appendix:cleaned_ecspell}
\begin{table}[tb!]
    \setlength{\tabcolsep}{4.5pt}
    \renewcommand{\arraystretch}{0.95}
    \centering
    \fontsize{9}{12}\selectfont
    \begin{NiceTabular}{llccccc}
        \toprule
        \rowcolor[gray]{1.0}
        \Block{2-1}{\textbf{Domain}} & \Block{2-1}{\textbf{Model}} & \Block{1-3}{\textbf{Correction}} & & & \Block{2-1}{$\mathrm{FPR}$} \\ 
        \cmidrule(lr){3-5}
        \rowcolor[gray]{1.0}
        & & $\mathrm{P}$ & $\mathrm{R}$ & $\mathrm{F}_1$ & \\ \midrule
        \rowcolor[gray]{.95}
        \Block{1-6}{ECSpell} & & & & & \\
        \Block{4-1}{LAW} & GPT3.5 & 48.5 & 43.1 & 45.6 & 9.4 \\
        & GPT4 & 62.0 & 62.0 & 62.0 & 7.3 \\
        & ReLM & 66.1 & \textbf{71.0} & 70.0 & 8.6 \\
        & \quad + DISC & \textbf{73.7} & 70.2 & \textbf{71.9} & \textbf{5.7} \\ \midrule
        \Block{4-1}{MED} & GPT3.5 & 36.5 & 42.0 & 39.1 & 20.1 \\
        & GPT4 & 45.1 & 57.5 & 50.6 & 24.8 \\
        & ReLM & 67.4 & 70.2 & 68.8 & 7.0 \\
        & \quad + DISC & \textbf{78.2} & \textbf{71.6} & \textbf{74.8} & \textbf{4.6} \\ \midrule
        \Block{4-1}{ODW} & GPT3.5 & 57.3 & 52.3 & 54.7 & 6.3 \\
        & GPT4 & 71.7 & 67.6 & 69.5 & \textbf{1.7} \\
        & ReLM & 78.2 & 76.4 & 77.3 & 4.1 \\
        & \quad + DISC & \textbf{81.8} & \textbf{76.7} & \textbf{79.2} & 2.5 \\
        \bottomrule
    \end{NiceTabular}
    \caption{Sentence-level performance of LLMs, ReLM, and
        ReLM + DISC on the test sets of cleaned ECSpell.}
    \label{tab:cleaned_ecspell}
\end{table}
We discovered a serious data leakage issue in ECSpell. Specifically, the same correct sentence may appear multiple times in both the training and test sets, with only the location and type of errors varying. These sentences respectively account for 52.7\%, 19.3\%, and 28.2\% of the ECSpell-LAW/MED/ODW training sets. We cleaned these duplicate sentences and reorganized the experiments, as shown in Table~\ref{tab:cleaned_ecspell}.

The experimental results show that, compared to using the uncleaned training sets, ReLM's performance on ECSpell significantly decreased, but it still outperforms GPT4. DISC also achieves stable performance improvement, with an impressive 6.0 F1 value increase on ECSpell-MED.

\section{Detailed Results}
\label{sec:appendix:detection_level results}
\begin{table*}[tb!]
    \renewcommand{\arraystretch}{0.95}
    \centering
    \scalebox{0.92}{
        \begin{NiceTabular}{lccc|ccc|ccc}
            \toprule
            \rowcolor[gray]{1.0}
            \Block[l]{2-1}{\textbf{Models}} & \Block[c]{1-3}{\textbf{SIGHAN15}} & & & \Block[c]{1-3}{\textbf{SIGHAN14}} & & & \Block[c]{1-3}{\textbf{SIGHAN13}} & & \\
            \rowcolor[gray]{1.0} & D-P\bgood & D-R\bgood & D-F\bgood & D-P\bgood & D-R\bgood & D-F\bgood & D-P\bgood & D-R\bgood & D-F\bgood \\
            \midrule
            \rowcolor[gray]{.95}
            \Block[c]{1-10}{{Previous SOTAs}} & & & & & & & & & \\
            \Block[l]{1-1}{SpellGCN} & 74.8 & 80.7 & 77.7 & 65.1 & 69.5 & 67.2 & 80.1 & 74.4 & 77.2 \\
            \Block[l]{1-1}{ReaLiSe} & 77.3 & 81.3 & 79.3 & 67.8 & 71.5 & 69.6 & 88.6 & 82.5 & 85.4 \\
            \Block[l]{1-1}{SCOPE\rlap{$^\dagger$}} & 80.5 & 85.4 & 82.9 & 68.8 & 73.7 & 71.1 & 87.5 & 83.0 & 85.2 \\
            \Block[l]{1-1}{SCOPE\;+\;DR-CSC} & \textbf{82.9} & 84.8 & \textbf{83.8} & \textbf{70.2} & 73.3 & 71.7 & 88.5 & 83.7 & 86.0 \\
            \Block[l]{1-1}{ReLM\rlap{$^\dagger$}} & 78.3 & \textbf{85.6} & 81.8 & 65.7 & 74.5 & 69.8 & 86.4 & 83.7 & 85.0 \\
            \midrule
            \rowcolor[gray]{.95}
            \Block[c]{1-10}{{LLMs Results}} & & & & & & & & & \\
            \Block[l]{1-1}{{GPT3.5}} & 39.4 & 46.4 & 42.6 & 41.4 & 23.1 & 29.6 & 61.6 & 29.2 & 39.7 \\
            \Block[l]{1-1}{{GPT4}} & 42.7 & 57.5 & 49.0 & 38.1 & 52.3 & 44.1 & 53.4 & 51.6 & 52.5 \\
            \midrule
            \rowcolor[gray]{.95}
            \Block[c]{1-10}{{Ours}} & & & & & & & & & \\
            \Block[l]{1-1}{ReaLiSe\;+\;DISC} & 78.3 & 81.2 & 79.7\rlap{$^\uparrow$} & 69.2 & 71.2 & 70.1\rlap{$^\uparrow$} & 88.9 & 82.2 & 85.4 \\
            \Block[l]{1-1}{SCOPE\;+\;DISC} & 81.7 & 84.8 & 83.2\rlap{$^\uparrow$} & \textbf{70.2} & 73.5 & 71.8\rlap{$^\uparrow$} & 88.8 & 83.7 & 86.2\rlap{$^\uparrow$} \\
            \Block[l]{1-1}{ReLM\;+\;DISC} & 80.8 & 84.3 & 82.5\rlap{$^\uparrow$} & 69.7 & \textbf{74.9} & \textbf{72.2}\rlap{$^\uparrow$} & \textbf{89.7} & \textbf{84.5} & \textbf{87.0}\rlap{$^\uparrow$} \\
            \bottomrule
        \end{NiceTabular}
    }
    \caption{
        Sentence-level performance on the SIGHAN13, SIGHAN14 and SIGHAN15 test sets. Precision ($\mathrm{P}$),
        recall ($\mathrm{R}$) and $\mathrm{F}_1$ for detection are reported (\%). Results marked with ``$\dagger$'' are obtained by reruning the official code released by \citet{Li-etal-2022-SCOPE} and \citet{Liu-etal-2024-ReLM}. Other baseline results are directly taken from their literature. Apart from SpellGCN, all models apply post-processing on SIGHAN13, which removes all detected and corrected ``地'' and ``得'' from the model output before evaluation. ``+ DISC'' means adding DISC module in the decoder. $\alpha$ and $\beta$ are assigned the values 1.1 and 0.7, respectively.
    }
    \label{tab:sighans_detection_level}
\end{table*}

In addition to the correction-level performance, we also present the detection-level experimental results of the CSC models, as shown in Table~\ref{tab:sighans_detection_level}. SCOPE + DR-CSC performs well at the detection level, primarily because they incorporate an additional detection network.

Since SIGHANs contain a lot of noise, we also conduct experiments on their revised versions (referred to as SIGHANs (rev.)) released by \citet{yang-etal-2023-chinese}, which have undergone manual verification and error correction to ensure higher data quality. As shown in Table~\ref{tab:rsighans_correction_level} and Table~\ref{tab:rsighans_detection_level}, our DISC module also achieves consistent performance improvements on SIGHANs (rev.).
\begin{table*}[tb!]
    \setlength{\tabcolsep}{3.85pt}
    \renewcommand{\arraystretch}{0.95}
    \centering
    \scalebox{0.92}{
        \begin{NiceTabular}{lcccc|cccc|cccc}
            \toprule
            \rowcolor[gray]{1.0}
            \Block[l]{2-1}{\textbf{Models}} & \Block[c]{1-4}{\textbf{SIGHAN15 (rev.)}} & & & & \Block[c]{1-4}{\textbf{SIGHAN14 (rev.)}} & & & & \Block[c]{1-4}{\textbf{SIGHAN13 (rev.)}} & & & \\
            \rowcolor[gray]{1.0} & C-P\bgood & C-R\bgood & C-F\bgood & FPR\sgood & C-P\bgood & C-R\bgood & C-F\bgood & FPR\sgood & C-P\bgood & C-R\bgood & C-F\bgood & FPR\sgood \\
            \midrule
            \rowcolor[gray]{.95}
            \Block[c]{1-13}{{Previous SOTAs}} & & & & & & & & & & & & \\
            \Block[l]{1-1}{BERT\rlap{$^\star$}} & 73.2 & 67.5 & 70.2 & -- & 62.6 & 57.5 & 59.9 & -- & 71.1 & 67.4 & 69.2 & -- \\
            \Block[l]{1-1}{ReaLiSe\rlap{$^\star$}} & 74.4 & 69.6 & 71.9 & -- & 63.6 & 59.0 & 61.2 & -- & 71.9 & 68.0 & 69.9 & -- \\
            \Block[l]{1-1}{\citet{yang-etal-2023-chinese}} & 77.0 & 67.6 & 72.0 & -- & 66.0 & 57.1 & 61.3 & -- & 73.2 & 67.1 & 70.0 & -- \\
            \Block[l]{1-1}{ReLM} & 76.4 & \textbf{73.5} & 74.9 & 8.5 & 65.5 & 62.9 & 64.2 & 11.3 & 74.0 & 70.9 & 72.4 & 10.7 \\
            \midrule
            \rowcolor[gray]{.95}
            \Block[c]{1-13}{{Ours}} & & & & & & & & & & & & \\
            \Block[l]{1-1}{ReLM\;+\;DISC} & \textbf{79.0} & 73.0 & \textbf{75.9} & \textbf{6.4} & \textbf{69.3} & \textbf{63.4} & \textbf{66.2} & \textbf{8.9} & \textbf{75.4} & \textbf{71.3} & \textbf{73.3} & \textbf{9.4} \\
            \bottomrule
        \end{NiceTabular}
    }
    \caption{
        Sentence-level performance on the revised SIGHAN13-15 test sets. Precision ($\mathrm{P}$),
        recall ($\mathrm{R}$) and $\mathrm{F}_1$ for correction are reported (\%). ``*'' means that the results of BERT and ReaLiSe in the table are directly copied from \citet{yang-etal-2023-chinese}.
    }
    \label{tab:rsighans_correction_level}
\end{table*}

\begin{table*}[tb!]
    \renewcommand{\arraystretch}{0.95}
    \centering
    \scalebox{0.92}{
        \begin{NiceTabular}{lccc|ccc|ccc}
            \toprule
            \rowcolor[gray]{1.0}
            \Block[l]{2-1}{\textbf{Models}} & \Block[c]{1-3}{\textbf{SIGHAN15 (rev.)}} & & & \Block[c]{1-3}{\textbf{SIGHAN14 (rev.)}} & & & \Block[c]{1-3}{\textbf{SIGHAN13 (rev.)}} & & \\
            \rowcolor[gray]{1.0} & D-P\bgood & D-R\bgood & D-F\bgood & D-P\bgood & D-R\bgood & D-F\bgood & D-P\bgood & D-R\bgood & D-F\bgood \\
            \midrule
            \rowcolor[gray]{.95}
            \Block[c]{1-10}{{Previous SOTAs}} & & & & & & & & & \\
            \Block[l]{1-1}{BERT\rlap{$^\star$}} & 75.4 & 70.0 & 72.4 & 64.6 & 59.3 & 61.8 & 72.6 & 68.8 & 70.6 \\
            \Block[l]{1-1}{ReaLiSe\rlap{$^\star$}} & 75.8 & 70.9 & 73.2 & 65.6 & 60.8 & 63.1 & 74.9 & 70.7 & 72.7 \\
            \Block[l]{1-1}{\citet{yang-etal-2023-chinese}} & 77.7 & 68.3 & 72.7 & 67.2 & 58.1 & 62.3 & 74.4 & 68.3 & 71.2 \\
            \Block[l]{1-1}{ReLM} & 78.8 & \textbf{75.7} & 77.2 & 68.4 & \textbf{65.7} & 67.0 & 76.0 & \textbf{72.8} & \textbf{74.4} \\
            \midrule
            \rowcolor[gray]{.95}
            \Block[c]{1-10}{{Ours}} & & & & & & & & & \\
            \Block[l]{1-1}{ReLM\;+\;DISC} & \textbf{80.6} & 74.4 & \textbf{77.4} & \textbf{71.2} & 65.2 & \textbf{68.1} & \textbf{76.4} & 72.3 & 74.3 \\
            \bottomrule
        \end{NiceTabular}
    }
    \caption{
        Sentence-level performance on the revised SIGHAN13-15 test sets. Precision ($\mathrm{P}$),
        recall ($\mathrm{R}$) and $\mathrm{F}_1$ for detection are reported (\%). ``*'' means that the results of BERT and ReaLiSe in the table are directly copied from \citet{yang-etal-2023-chinese}.
    }
    \label{tab:rsighans_detection_level}
\end{table*}

\end{CJK*}
\end{document}